\documentclass[lettersize,journal]{IEEEtran}
\usepackage{amsmath,amsfonts}
\usepackage{algorithmic}
\usepackage{algorithm}
\usepackage{array}
\usepackage[caption=false,font=normalsize,labelfont=sf,textfont=sf]{subfig}
\usepackage{textcomp}
\usepackage{stfloats}
\usepackage{url}
\usepackage{verbatim}
\usepackage{graphicx}
\usepackage{cite}
\graphicspath{{figures/}}
\usepackage[colorlinks=true,linkcolor=blue,urlcolor=blue,citecolor=blue]{hyperref}
\makeatletter
\def\hlinew#1{%
  \noalign{\ifnum0=`}\fi\hrule \@height #1 \futurelet
   \reserved@a\@xhline}
\usepackage{multirow}
\usepackage{pifont}
\newcommand{\cmark}{\ding{51}}
\newcommand{\xmark}{\ding{55}}
\newcommand{\tcr}{\textcolor{red}}
\newcommand{\tcg}{\textcolor{cyan}}

\begin{document}

\title{SpectralMamba: Efficient Mamba for Hyperspectral Image Classification}

\author{Jing Yao,~\IEEEmembership{Member,~IEEE,}
        Danfeng Hong,~\IEEEmembership{Senior Member,~IEEE,}
        Chenyu Li,
        and Jocelyn Chanussot,~\IEEEmembership{Fellow,~IEEE}

\thanks{This work was supported by the National Natural Science Foundation of China under Grant 62201553, 42241109, and 42271350.} 
%(\emph{Corresponding author: Danfeng Hong})}
\thanks{J. Yao is with the Aerospace Information Research Institute, Chinese Academy of Sciences, Beijing 100094, China. (e-mail: yaojing@aircas.ac.cn)}
\thanks{D. Hong is with the Aerospace Information Research Institute, Chinese Academy of Sciences, 100094 Beijing, China, and also with the School of Electronic, Electrical and Communication Engineering, University of Chinese Academy of Sciences, Beijing 100094, China. (e-mail: hongdf@aircas.ac.cn).}
\thanks{C. Li is with the Aerospace Information Research Institute, Chinese Academy of Sciences, 100094 Beijing, China, and also with the School of Mathematics and Statistics, Southeast University, 211189 Nanjing, China. (e-mail: lichenyu@seu.edu.cn)}
\thanks{J. Chanussot is with Univ. Grenoble Alpes, Inria, CNRS, Grenoble INP, LJK, Grenoble 38000, France, and also with the Aerospace Information Research Institute, Chinese Academy of Sciences, Beijing 100094, China. (e-mail: jocelyn.chanussot@inria.fr)}
%\thanks{Z. Xu is with the School of Mathematics and Statistics, Xi’an Jiaotong University, Xi’an 710049, China. (email: zbxu@mail.xjtu.edu.cn)}  
}

\markboth{}%
{Shell \MakeLowercase{\textit{et al.}}: A Sample Article Using IEEEtran.cls for IEEE Journals}


\maketitle

\begin{abstract}
% How to effectively characterize the sequential properties among hundreds of bands for hyperspectral data has long been recognized as a crucial research topic. 
Recurrent neural networks and Transformers have recently dominated most applications in hyperspectral (HS) imaging, owing to their capability to capture long-range dependencies from spectrum sequences.
However, despite the success of these sequential architectures, the non-ignorable inefficiency caused by either difficulty in parallelization or computationally prohibitive attention still hinders their practicality, especially for large-scale observation in remote sensing scenarios. To address this issue, we herein propose SpectralMamba -- a novel state space model incorporated efficient deep learning framework for HS image classification.
SpectralMamba features the simplified but adequate modeling of HS data dynamics at two levels.
First, in spatial-spectral space, a dynamical mask is learned by efficient convolutions to simultaneously encode spatial regularity and spectral peculiarity, thus attenuating the spectral variability and confusion in discriminative representation learning.
Second, the merged spectrum can then be efficiently operated in the hidden state space with all parameters learned input-dependent, yielding selectively focused responses without reliance on redundant attention or imparallelizable recurrence. To explore the room for further computational downsizing, a piece-wise scanning mechanism is employed in-between, transferring approximately continuous spectrum into sequences with squeezed length while maintaining short- and long-term contextual profiles among hundreds of bands.
Through extensive experiments on four benchmark HS datasets acquired by satellite-, aircraft-, and UAV-borne imagers, SpectralMamba surprisingly creates promising win-wins from both performance and efficiency perspectives. The code will be available at \url{https://github.com/danfenghong/SpectralMamba} for the sake of reproducibility.
\end{abstract}


\section{Introduction}
% \IEEEPARstart{S}{ince} its advent, hyperspectral (HS) imaging has broken through the limitations of conventional photography by simultaneously capturing both spatial and spectral information, allowing humans to 
% enabling humans a more detailed and profound observation of the real world. % surpasses

\IEEEPARstart{T}{he} emergent development of hyperspectral (HS) imaging remarkably empowers humans in observing the real world in greater detail and depth \cite{bioucas2013hyperspectral}.
Unlike traditional photography which acquires images in a limited number of broad spectral bands, the HS imaging system unprecedentedly achieves the simultaneous capture of both spatial and spectral information by measuring a spectrum of the energy at each pixel.
The generated 3-dimensional (3-D) HS data cube contains the near-continuous spectral profile for each spatial resolution element, thus allowing for more accurate quantification, identification, and recognition of the imaged contents.
Owing to the recent progress in aerospace and instrumental technology \cite{xi2022few}, HS imaging has gradually become an indispensable tool for remote sensing (RS).
Of all the widespread applications it has found, HS image classification has drawn considerable attention across diverse fields, from environmental monitoring, urban planning, to military science, demonstrating its pervasive potential and cross-cutting importance \cite{hong2024spectralgpt,chang2023changes}.

\begin{figure}[!t]
    \centering
		\includegraphics[width=0.48\textwidth]{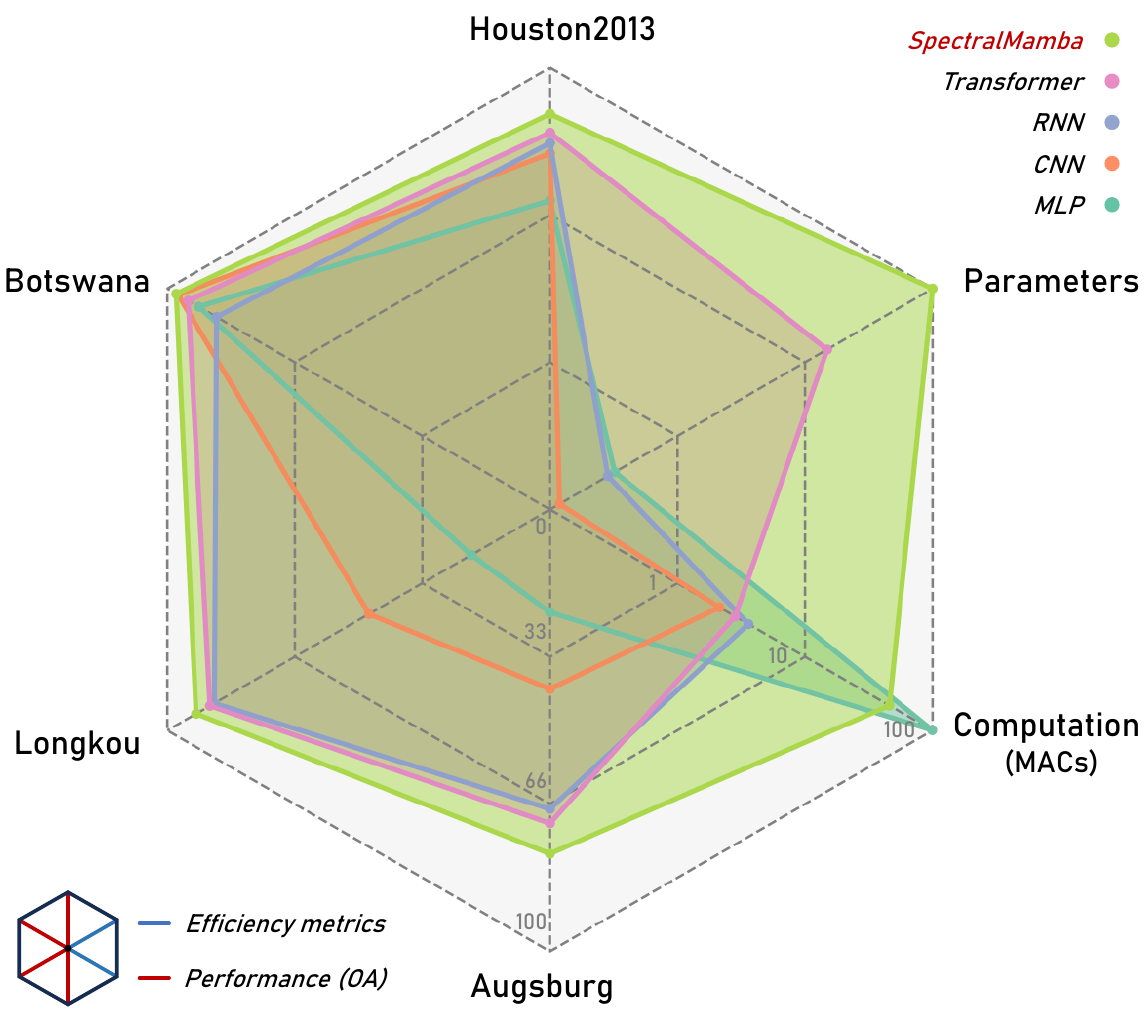}
    \caption{The radar chart of experimental results of SpectralMamba and classic network architectures in terms of both hyperspectral image classification performance metric (OA) and mean efficiency metrics (number of parameters and MACs) on four benchmark datasets.
     To better visualize their differences, we set the lowest values of parameter number and MACs as the base score of 100, and customize a base-10 logarithmic scale on the MACs-axis.
    %  According to the chart, our SpectralMamba successfully achieves the maximum area by significantly surpassing its competitors along most metrics, showcasing its great potential as a novel efficient and effective DL framework for HS data analysis.
    According to the chart, our SpectralMamba significantly outperforms its competitors along most metrics, showcasing its great potential as a novel efficient, and effective deep learning framework for hyperspectral data analysis.
     }
    \label{fig:radar}
\end{figure}

The primary objective of HS image classification in RS is to accurately identify the various land cover or land use types of interest within the image by leveraging the detailed spectral signatures associated with each pixel \cite{ahmad2022hyperspectral}.
Despite the capability of HS imaging to capture hundreds of narrow wavelength bands (typically ranging from the visible spectrum to the near-infrared region), providing in-depth characterization of the spectral properties of different materials, there remain two long-standing challenges in its practical applications.

\begin{enumerate}
    \item The curse of dimensionality, also known as the Hughes phenomenon, is often encountered in HS image classification \cite{hong2021joint} when the classification accuracy undergoes an initial rise with more spectral bands observed but then decreases dramatically once a certain number of bands is reached \cite{ma2013hughes}. The fundamental cause of this issue stems from the exponential growth of the feature space volume as the number of dimensions expands, rendering the computational processing and effective analysis of the HS data increasingly burdensome and challenging.
    \item Spectral variability and spectral confusion are the other two prevalent phenomena that frequently manifest in HS data.
    The former issue refers to the situation where the same material displays varying spectral characteristics under different conditions, such as changes in illumination, atmospheric effects, or intrinsic variations, while the latter occurs when distinct materials exhibit similar spectral profiles \cite{theiler2019spectral}.
\end{enumerate}
In addition, several other issues always arise in conjunction with these challenges in HS data analysis, such as the limited availability of labeled training samples, and inevitable sensor noise with complex distributions, which makes it more challenging to precisely distinguish ground objects based solely on their spectral reflectances.
% which obviously precludes the direct applicability of distinguishing ground objects 

To tackle these challenges, researchers have dedicated significant efforts over the past decades to develop ever-advancing dimensionality reduction and feature extraction techniques for precise pixelwise recognition of HS images.
In the early stages, researchers investigated the applicability of statistical approaches like principal component analysis, independent component analysis, kernel methods \cite{camps2005kernel}, and linear discriminant analysis \cite{camps2013advances}, as well as machine learning and heuristics techniques encompassing subspace learning \cite{hong2021multimodal}, manifold learning \cite{hong2019learnable}, ensemble methods \cite{samat2014rm}, and active learning strategies \cite{haut2018active}, to effectively process and analyze HS data.
During this period, shallow machine learning models, such as nearest neighbors, decision trees, and support vector machines, emerged as prevalent choices for effective backend classifiers complementing these feature extraction methods.

As deep learning (DL) has proliferated across numerous research fields since the last decade, the RS community has also embraced this powerful learning paradigm for HS data analysis, harnessing its capability to learn representations directly from the data, thereby mitigating the intrinsic cognitive bias imposed by inadequate mathematical modeling in conventional approaches \cite{li2023lrr,xu2023txt2img}.
Among the various DL architectures for HS image classification, convolution neural network (CNN) has taken pride of place over a long period. 
Benefiting from the local receptive ability through shift-invariant convolutions, typical CNNs successfully realize hierarchical feature extraction and semantic abstraction in end-to-end training from input-target pairs.
However, although this family of models is good at exploiting local contextual information, their inherent local connectivity and weight sharing inevitably restrict the modeling of long-range correlations and dynamics within and across data sequences, respectively.

\begin{figure*}[!t]
    \centering
		\includegraphics[width=1\textwidth]{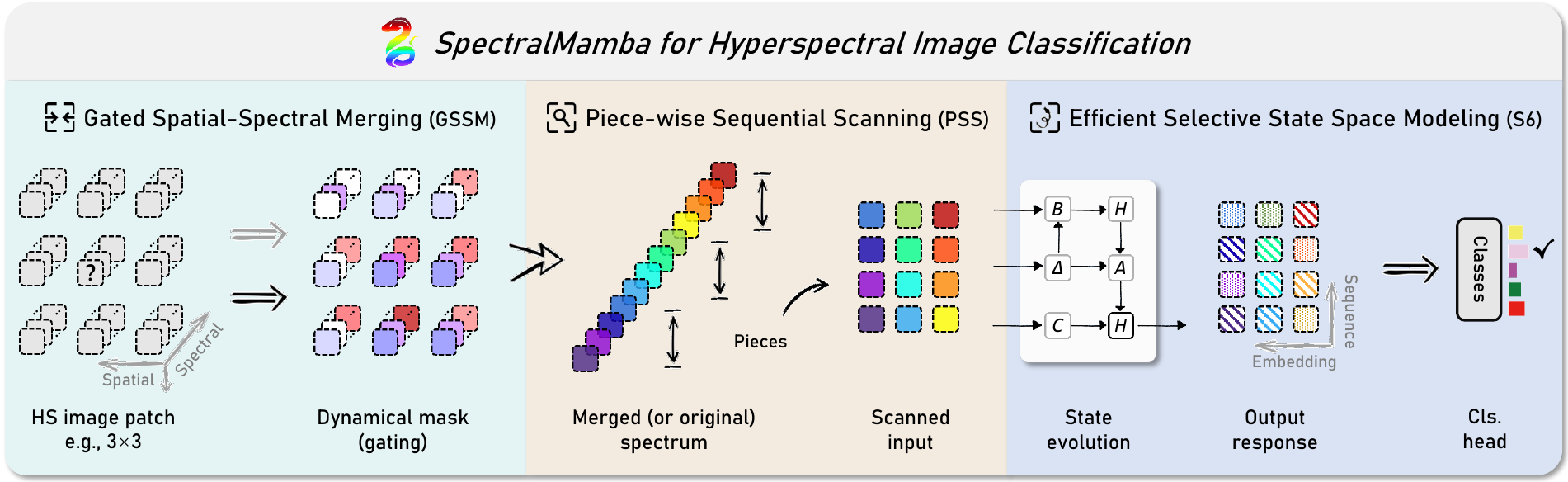}
    \caption{SpectralMamba mainly consists of three components, i.e., gated spatial-spectral merging (GSSM) module, piece-wise sequential scanning (PSS) strategy, and efficient selective state space (S6) modeling.
    Its pixelwise counterpart functions by directly operating the original spectrum from the middle stage.
    $A$, $B$, $C$, and $\Delta$ denote all learnable parameters in the hidden state space, where $H$ records the selectively embedded sequence for the final output.
    }
    \label{fig:workflow}
\end{figure*}

At this point, sequence models, such as recurrent neural networks (RNNs) and Transformers, came into the public eye for their effectiveness in processing sequential data.
By orderly unfolding the HS spectrum into a long sequence, RNNs and Transformers essentially enable the capture of long- and short-term spectral fingerprints through recurrent state modeling and attention mechanism, respectively.
Besides, these sequence models have been widely proven to be more competent in handling non-linear data dynamics under complex RS scenes than CNNs.
Despite these merits, they inherently suffer from parallel training difficulty and burdensome pairwise multiplicative computations, respectively.
Although enormous efforts have been made to cope with these issues, as of yet,  most of these variants can hardly avoid becoming increasingly sophisticated in either network structures or working flows in pursuing accuracy breakthroughs with improved representation, appearing to have reached a plateau that cannot avoid trading-off between the performance and computational efficiency.

% hampering the accuracy and reliability of the classification results.

% Differ

% cost-effective and computation-friendly

% achieved favorable results 
Fortunately, recent advances in the state space model (SSM) make it broadly applicable and provide a novel avenue for sequentiality modeling.
Building upon the theoretical foundation of Classical SSM from control theory and powerful modern DL advantages, the emerging deep SSMs unprecedentedly allow for efficient computation in learning very long-range dependencies over tens of thousands of timesteps and are dominating more and more benchmarks across various domains \cite{gu2021efficiently}.
Nevertheless, existing SSMs are typically designed for causal learning on low-dimensional sequences such as audio and language, leaving their practicality on high-dimensional visual data such as HS imagery underexplored.
Therefore, in this work, we excavate the potential of tailoring SSM to HS data by taking a deep dive into their traits.
To be more specific, we propose SpectralMamba -- an efficient and effective SSM-integrated DL framework for both pixelwise and patchwise input-based HS image classification.
SpectralMamba leverages a simplified but adequate modeling of HS data dynamics in both spatial-spectral feature space and hidden state space, thereby alleviating the effect caused by spectral variability and spectral confusion.
The underlying computational overhead caused by parameter size and computations is further reduced by a customized scanning strategy that can additionally enhance sequential representation while maintaining local spectral fingerprints of HS data.
The main contributions of this article can be highlighted as follows.
\begin{enumerate}
    \item We propose a novel SSM-based backbone network termed SpectralMamba that makes a further step towards performant and computation-friendly HS image classification from the perspective of sequence modeling.
    To the best of our knowledge, this is the first work that well tailors the deep SSM for HS data and its analysis.
    \item Targeting the high dimensionality of HS data, and the issues of spectral variability and confusion, we propose the strategies of piece-wise sequential scanning (PSS) and gated spatial-spectral merging (GSSM) to fully encode the underlying spatial regularity and spectral peculiarity, yielding a more robust discriminative representation learned via a fully lightweight architecture.
    \item Through extensive experimental comparison on four benchmark HS datasets acquired from satellite-, aircraft-, and UAV-borne platforms, our SpectralMamba significantly outperforms the representative competitors with classic backbones at a generally minimal computational resource cost (shown in Fig. \ref{fig:radar}).
    The ablation studies further verify the effectivity of our key components, such as PSS maximally brings approximately 4$\%$ improvement in OA while reducing 60$\%$ parameters and 40$\%$ computations than our baseline.
\end{enumerate}
% a stable performance at a high degree of performance.

The remainder of this article is organized as follows. 
Section \ref{S2} introduces the preliminary elements for state space models, elaboration of our SpectralMamba, and method analysis.
Section \ref{S3} details the experiments, including descriptions of the datasets and implementation, evaluation of both performance and computational cost, comparison results and analysis, and ablation studies.
Finally, Section \ref{S4} concludes the work and points out plausible future directions.

% radar figure

% An Illustration to clarify how our proposed SpectralMamba works, its detailed network architecture and data flow in the process for hyperspectral image classification.

\section{Methodology}\label{S2}

\subsection{Preliminaries}
\subsubsection{State Space Model} Inspired from classical SSMs \cite{kalman1960new} and modern DL advances, most notably CNNs, RNNs, and Transformers, structured state space sequence models (S4) \cite{gu2021efficiently, nguyen2022s4nd} have recently emerged and garnered considerable attention for modeling sequential data.
This class of models typically originates from a continuous-time system that maps an input function or sequence $x(t)\in\mathbb{R}^M$ to an output response signal $y(t)\in\mathbb{R}^O$ through an implicit latent state $h(t)\in\mathbb{R}^N$, which can be mathematically formulated using the following ordinary differential equations,
\begin{align}
    h'(t)&=\mathbf{A}h(t)+\mathbf{B}x(t),\\
    y(t)&=\mathbf{C}h(t)+\mathbf{D}x(t),
\end{align}
where $\mathbf{A}\in\mathbb{R}^{N\times N}$ and $\mathbf{C}\in\mathbb{R}^{O\times N}$ control how the current state evolves over time and translates to the output, $\mathbf{B}\in\mathbb{R}^{N\times M}$ and $\mathbf{D}\in\mathbb{R}^{O\times M}$ depict how the input influences the state and the output, respectively.
Herein we consider the case of a single-input single-output system with $O=M=1$ and omit $\mathbf{D}x(t)$ term by treating it as a skip connection as the S4 model does \cite{gu2021efficiently}.
% Mathematically, it functions as a continuous linear time-invariant system which .

\subsubsection{Discretization} 
The first step in applying SSMs to discrete signals such as language, audio, and images, is to transform the system parameters into their ``discretized" counterpart.  
A commonly adopted discretization method is the zero-order hold rule, by which the reparameterization holds as follows,
\begin{align}
    \mathbf{\bar{A}}&=\exp(\Delta\mathbf{A}),\\
    \mathbf{\bar{B}}&=(\Delta\mathbf{A})^{-1}(\mathbf{\bar{A}}-\mathbf{I})(\Delta\mathbf{B})\nonumber\\
    &\approx(\Delta\mathbf{A})^{-1}(\Delta\mathbf{A})(\Delta\mathbf{B})\\
    &=\Delta\mathbf{B},\nonumber
\end{align}
where the first-order Taylor series approximation is used by following \cite{gu2023mamba}. 
The timescale parameter $\Delta$ denotes the sampling step, i.e., $x_k=x(k\Delta)$, which also trades off the state and current input during the evolution.
Then the discrete SSM can be formulated into the following recurrent representation,
\begin{align}
    h_k&=\mathbf{\bar{A}}h_{k-1}+\mathbf{\bar{B}}x_k,\\
    y_k&=\mathbf{C}h_k,
\end{align}
which can be computed similarly to RNNs.
To better accommodate GPU acceleration for efficient training, S4 also unrolls the above linear recurrence, yielding its global convolutional representation as
\begin{equation}
    \mathbf{y}=\mathbf{x}*\mathbf{\bar{K}},
\end{equation}
where $\mathbf{\bar{K}}=(\mathbf{C}\mathbf{\bar{B}},\mathbf{C}\mathbf{\bar{A}}\mathbf{\bar{B}},\ldots,\mathbf{C}\mathbf{\bar{A}}^{L-1}\mathbf{\bar{B}})$ represents the SSM convolution kernel, and $L$ is the sequence length of input.

\subsubsection{Mamba} 
Besides the linearity, another simplifying assumption of the above systems is the time invariance, that is, all the system parameters are defined as time-independent. 
Recently, a novel class of selective SSMs (S6) breaks this constraint by parameterizing $(\Delta,\mathbf{B},\mathbf{C})$ as functions of input $\mathbf{x}$, thus endowing SSMs with additional selection ability to focus on the important or ignore the unimportant.
% thus enabling SSMs to select import information in a content-aware manner.
As a common fashion, a simplified neural network architecture consists of linear layers, a convolution layer, a residual connection, nonlinear transformations, and most importantly the S6 kernel together to form the Mamba block.
A hardware-aware optimization is also proposed to guarantee its efficient implementation.

\begin{figure*}[!t]
    \centering
		\includegraphics[width=1\textwidth]{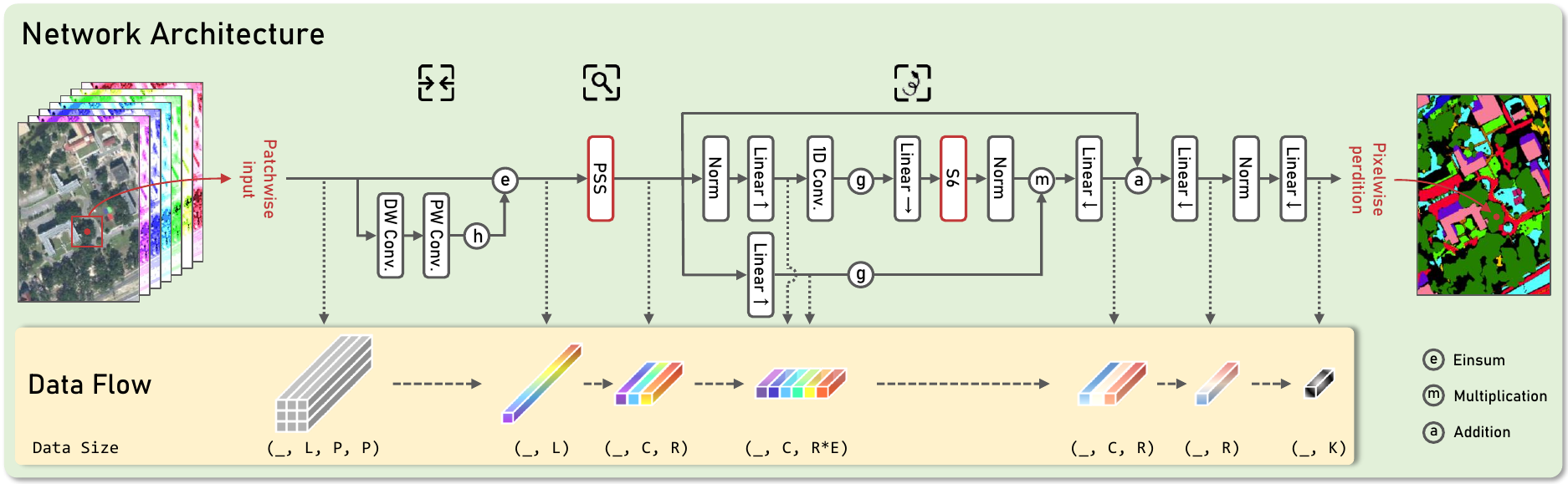}
    \caption{Detailed architectural design and data processing pipeline of our proposed SpectralMamba, exemplified using a patchwise input under batch training for hyperspectral image classification, where PSS and S6 refer to the piece-wise sequential scanning and selective state space model, respectively.}
    \label{fig:network}
\end{figure*}

\subsection{SpectralMamba: Overview}
To break through the performance and efficiency bottlenecks of existing methods based on CNN, RNN, or Transformer backbones, we propose the SpectralMamba, a Mamba -- S6 model incorporated DL solution to tackle HS image classification.
The key of SpectralMamba lies in its simultaneous modeling of HS data dynamics by gated spectrum merging in the spatial-spectral space and selective sequence learning in hidden state space through a minimally parameterized network architecture.
Furthermore, a novel sequential scanning strategy tailored for HS data is proposed to make the framework more computation-friendly by unraveling the hundreds-of-bands spectrum into pieces.
The patchwise SpectralMamba can also be flexibly transformed into its pixelwise counterpart by skipping the spatial-spectral encoding stem.
Fig. \ref{fig:workflow} gives an illustration of how our proposed SpectralMamba works, and the detailed network architecture as well as its data flow are shown in Fig. \ref{fig:network}.

\subsection{SpectralMamba: Key Components}
Let the 1-D vector $\mathbf{x}^{pixel}=[\mathbf{x}^{pixel}_1,\ldots,\mathbf{x}^{pixel}_L]\in\mathbb{R}^{1\times L}$ denote a given pixel of an HS image, where $L$ is the band number of the spectrum.
We consider the state space modeling within the spectral domain, that is, our aim becomes to find its output response $\mathbf{y}^{pixel}\in\mathbb{R}^L$ through a well-defined S6 model.
However, simply treating the reflectance values at each band as the representation may limit the mining of sequential patterns.
Therefore we expand the model dimension by a factor, $E=8$ in our case, to enlarge the state space size.
As the core architectural part in Fig. \ref{fig:network} depicts, the Mamba block designed for HS data consists of three streams.
Its mainstream comprises two distinct LayerNorm layers in input and expanded state spaces, three linear layers for expanding, keeping, and compressing feature dimensions, a SiLU nonlinear activation, and the S6 block.
The other two streams are the skip connection and an excitation-like multiplication to adaptively transform original information across layers \cite{hu2018squeeze}.
Note that the use of skip connection and nonlinearity is crucial for stable training with fast convergence, while the practical performance appears to be not that sensitive to the choice of normalization and activation function.

\subsubsection{Piece-wise Sequential Scanning}
The aforementioned state space modeling is empowered to pay attention or forget features at particular wavelengths through an input-dependent parameterizing way \cite{gu2023mamba}.
When applied to HS data with hundreds of near-continuous bands, its high spectral redundancy drives us to rethink the input manner.
Unlike the very recent efforts in modifying Mamba for natural image processing by considering spatial multi-directional scanning \cite{liu2024vmamba}, we propose a novel piece-wise sequential scanning (PSS) along the spectral dimension to fully leverage the reflectance characteristics of different types of ground objects.

In specific, we can formulate the PSS module as \begin{equation}
    PSS(\mathbf{x}^{pixel})=[S_1\mathbf{x}^{pixel},\ldots,S_R\mathbf{x}^{pixel}],\label{eq:PSS}
\end{equation}
where $S_r\mathbf{x}^{pixel}=[\mathbf{x}^{pixel}_{(r-1)C+1},\ldots,\mathbf{x}^{pixel}_{rC}]^\top\in\mathbb{R}^{C\times1}$ scans continuous pieces from the original spectrum $\mathbf{x}^{pixel}$ for $r=1,\ldots,R$, and $R$ is the number of pieces with length $C$.
This also acts similarly as a resampling with features at each position in the sequence enriched.
By applying PSS before our Mamba block, the response accordingly turns from a 1-D $L$-length vector to a 2-D output of shape $C\times R$.
We then add one more pre-layer before the common softmax-based classification head to finally obtain $K$-length categorical logits.

\subsubsection{Gated Spatial-Spectral Merging}
It is crucially important to consider spatial information for discriminative representation learning.
However, conventional spatial-spectral feature extraction methods commonly treat each patch equally with fixed convolution kernels.
Inspired by the S6 model that learns interactions along the sequence in an input-dependent way, we propose to further increase the content-awareness by introducing a dynamical gate function for adaptive spatial-spectral embedding.
With the proposed gated spatial-spectral merging (GSSM), we can substitute the $\mathbf{x}^{pixel}$ in Eq. (\ref{eq:PSS}) with merged spectrum computed as follows,
\begin{equation}
    GSSM(\mathbf{x}^{patch})=h(f_{PW}(f_{DW}(\mathbf{x}^{patch})))\otimes\mathbf{x}^{patch},
\end{equation}
where $h$ is the sigmoid activation, $f$ represents the composite function composed of depthwise (DW) convolution and pointwise (PW) convolution, $\otimes$ denotes the Einstein summation along the spatial dimension that combines two tensors of shape $L\times P\times P$ into a 1-D vector with length $L$.
Through GSSM, we hope to adaptively encode the semantic relationships between the center pixel and its neighborhoods in learning a more discriminative ``spectrum", thereby attenuating spectral variability and confusion effect.

\begin{table*}[!t]
% \scriptsize
\centering
\caption{Summary of four investigated HS datasets, including the tags of data acquisition information, categorical information of ground objects, and the corresponding numbers of train and test samples.}
% \vspace{2mm}
\resizebox{1\textwidth}{!}{ 
\begin{tabular}{c||ccc|ccc|ccc|ccc}
\hlinew{1pt}
\bf{Dataset}	&	\multicolumn{3}{c|}{\bf{Houston2013}}					&	\multicolumn{3}{c|}{\bf{Augsburg}}					&	\multicolumn{3}{c|}{\bf{Longkou}}					&	\multicolumn{3}{c}{\bf{Botswana}}					\\
\hline\hline																									
Sensor	&	\multicolumn{3}{c|}{ITRES CASI-1500}					&	\multicolumn{3}{c|}{HySpex}					&	\multicolumn{3}{c|}{Headwall Nano-Hyperspec}					&	\multicolumn{3}{c}{Hyperion}					\\
Platform	&	\multicolumn{3}{c|}{Aircraft-borne}					&	\multicolumn{3}{c|}{Aircraft-borne}					&	\multicolumn{3}{c|}{UAV-borne}					&	\multicolumn{3}{c}{Satellite-borne}					\\
Loc. \& Time	&	\multicolumn{3}{c|}{America, 2012}					&	\multicolumn{3}{c|}{Europe, 2018}					&	\multicolumn{3}{c|}{Asia, 2018}					&	\multicolumn{3}{c}{Africa, 2001}					\\
\hline
GSD	&	\multicolumn{3}{c|}{10 m}					&	\multicolumn{3}{c|}{30 m}					&	\multicolumn{3}{c|}{0.463 m}					&	\multicolumn{3}{c}{30 m}					\\
Wavelength	&	\multicolumn{3}{c|}{380 nm - 1050 nm}					&	\multicolumn{3}{c|}{400 nm - 2500 nm}					&	\multicolumn{3}{c|}{400 nm - 1000 nm}					&	\multicolumn{3}{c}{400 nm - 2500 nm}					\\
Data Size	&	\multicolumn{3}{c|}{349 $\times$ 1905 $\times$ 144}					&	\multicolumn{3}{c|}{332 $\times$ 485 $\times$ 180}					&	\multicolumn{3}{c|}{550 $\times$ 400 $\times$ 270}					&	\multicolumn{3}{c}{1476 $\times$ 256 $\times$ 145}					\\
\hline\hline							
Class No.	&	Class Name	&	Train	&	Test	&	Class Name	&	Train	&	Test	&	Class Name	&	Train	&	Test	&	Class Name	&	Train	&	Test	\\
\hline									
1	&	Healthy Grass	&	198	&	1053	&	Forest	&	80	&	13427	&	Corn	&	54	&	34457	&	Water	&	24	&	246	\\
2	&	Stressed Grass	&	190	&	1064	&	Residential Area	&	80	&	30249	&	Cotton	&	53	&	8321	&	Hippo Grass	&	20	&	81	\\
3	&	Synthetic Grass	&	192	&	505	&	Industrial Area	&	80	&	3771	&	Sesame	&	54	&	2977	&	Floodplain Grasses 1	&	21	&	230	\\
4	&	Tree	&	188	&	1056	&	Low Plants	&	80	&	26777	&	Broad-Leaf Soybean	&	53	&	63159	&	Floodplain Grasses 2	&	24	&	191	\\
5	&	Soil	&	186	&	1056	&	Allotment	&	80	&	495	&	Narrow-Leaf Soybean	&	54	&	4097	&	Reeds	&	20	&	249	\\
6	&	Water	&	182	&	143	&	Commercial Area	&	80	&	1565	&	Rice	&	50	&	11804	&	Riparian	&	20	&	249	\\
7	&	Residential	&	196	&	1072	&	Water	&	81	&	1449	&	Water	&	53	&	67003	&	Fires Car	&	20	&	239	\\
8	&	Commercial	&	191	&	1053	&	-	&	-	&	-	&	Roads and Houses	&	53	&	7071	&	Island Interior	&	20	&	183	\\
9	&	Road	&	193	&	1059	&	-	&	-	&	-	&	Mixed Weed	&	51	&	5178	&	Acacia Woodlands	&	21	&	293	\\
10	&	Highway	&	191	&	1036	&	-	&	-	&	-	&	-	&	-	&	-	&	Acacia Shrub Lands	&	22	&	226	\\
11	&	Railway	&	181	&	1054	&	-	&	-	&	-	&	-	&	-	&	-	&	Acacia Grasslands	&	20	&	285	\\
12	&	Parking Lot 1	&	192	&	1041	&	-	&	-	&	-	&	-	&	-	&	-	&	Short Mopane	&	22	&	159	\\
13	&	Parking Lot 2	&	184	&	285	&	-	&	-	&	-	&	-	&	-	&	-	&	Mixed Mopane	&	20	&	248	\\
14	&	Tennis Court	&	181	&	247	&	-	&	-	&	-	&	-	&	-	&	-	&	Exposes Soils	&	20	&	75	\\
15	&	Running Track	&	187	&	473	&	-	&	-	&	-	&	-	&	-	&	-	&	-	&	-	&	-	\\
\hline\hline			
Total	&	-	&	2832	&	12197	&	-	&	561	&	77733	&	-	&	475	&	204067	&	-	&	294	&	2954	\\
\hlinew{1pt}
\end{tabular}
}
\label{tab:datasets}
\end{table*}

\subsection{SpectralMamba: Method Analysis}
% ,  are adequately and efficiently captured 
% Specifically, by ably introducing the latest selective SMM
% one of the non-ignorable issues still has found its lies in their computational inefficiency in either training phase or inference.

It is non-trivial to directly extend SSM to applications of HS data.
Building upon the insights on SSM and structural prior knowledge of HS data, our SpectralMamba offers a viable SSM-based baseline to address the dense prediction application of HS images.
The proposed PSS strategy not only enables the model to uncover local characteristics of the spectral profile but also further improves efficiency by narrowing the width of core operation networks.
Moreover, the GSSM module is designed based on the observation that the semantic relationships between the central pixel and its neighboring pixels usually vary spatially and spectrally across the scene.
The widely-existed mixed pixel phenomenon can also differ for pixels within a local patch, especially for those falling on boundaries.
We hope to efficiently capture such highly spatial-spectral-variant HS data dynamics through a lightweight mask learner, thus yielding a merged spectrum with increased discriminative ability for the subsequent sequential learning in the state space.

The connections between our proposed SpectralMamba and related works are also worth noting.
On one hand, although CasRNN -- an RNN-based representative for HS image classification -- has considered the similar spectral redundancy issue by hierarchically learning from adjacent spectral bands to nonadjacent ones, their imparallelizable recurrence trait still accumulated both the computations and parameters in pursuing a stable training \cite{hang2019cascaded}.
Also, the unbearable quadratic complexity of self-attention in conventional Transformers has significantly magnified the computational burden as the number of neighboring spectral bands considered in a so-called groupwise embedding increases \cite{hong2022spectralformer}.
In contrast, the proposed PSS in our SpectralMamba perfectly matches the efficient feature selection via state evolution in S6, simultaneously preserving local spectral patterns and enlarging the feature dimension of sequence, while further improving computational efficiency.
What's more, our practice verifies that a non-overlapped scanning, i.e., $R=L/C$, is enough to produce a promising performance at a lower computational overhead.
On the other hand, different from conventional gated convolutions for natural images \cite{yu2019free}, our GSSM provides a lightweight gating mechanism in capturing the highly spatial-spectral-variant HS data dynamics.
It not only fits the setting to maintain the spectral sequentiality but also complements the subsequent content-aware learning in the state space with efficient spatial-spectral rectification.
In the following experiments section, we will demonstrate how SpectralMamba outperforms these predecessors by enhancing the efficacy in interpreting HS data on the abovementioned merits, meanwhile, maintaining high efficiency with low computational resource requirements.
% This is achieved by building upon the foundations laid by earlier work in this field.

\section{Experiments}\label{S3}
\subsection{Datasets}

We select four benchmark HS datasets to conduct experiments, covering all types of acquisition platforms, that are, aircraft-borne, unmanned aerial vehicle (UAV)-borne, and satellite-borne, hoping to give a comprehensive and faithful verification of our proposed SpectralMamba. 
The details are summarized in Table \ref{tab:datasets}.

\subsubsection{Houston2013 HS Dataset}

The first HS dataset Houston2013 consists of HS imagery acquired by ITRES CASI-1500 HS imager with 144 spectral bands ranging from 380 nm to 1050 nm. 
The investigated scene has 349 $\times$ 1905 pixels at a ground sampling distance (GSD) of 10 m and 15 LULC categories, covering the University of Houston campus and the surrounding urban area. 
It was provided by the IEEE Geoscience and Remote Sensing Society data fusion contest in 2013 and has been widely used in research and competitions related to HS image analysis and pixel-based classification \cite{yao2023extended}.

\begin{figure*}[!t]
    \centering
		\includegraphics[width=0.99\textwidth]{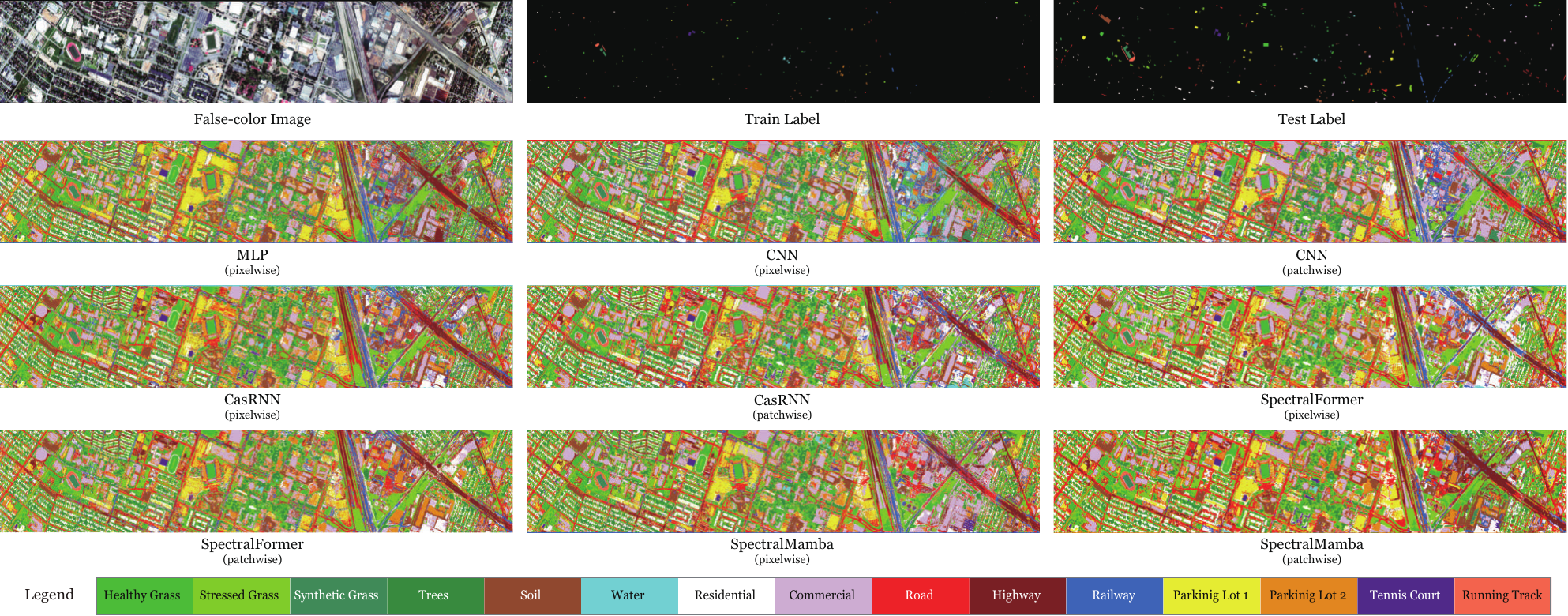}
    \caption{An illustration of the false-color image, train and test labels, and classification maps obtained by compared methods on the Houston2013 HS dataset.}
    \label{fig:HU}
\end{figure*}

\begin{table*}[!t]
\scriptsize
\centering
\caption{Quantitative comparison with related representatives on Houston2013 HS dataset. 
The best and second-best overall results are shown in bold and underlined, respectively.}
% \resizebox{0.99\textwidth}{!}{
\begin{tabular}{c||c|cc|cc|cc|cc}
\hlinew{1pt}
Method	&	MLP\cite{hong2023cross}	&	\multicolumn{2}{c|}{CNN\cite{yao2022semi}}			&	\multicolumn{2}{c|}{CasRNN\cite{hang2019cascaded}}			&	\multicolumn{2}{c|}{SpectralFormer\cite{hong2022spectralformer}}			&	\multicolumn{2}{c}{SpectralMamba}			\\
\hline																			
Implementation	&	Pixelwise	&	Pixelwise	&	Patchwise	&	Pixelwise	&	Patchwise	&	Pixelwise	&	Patchwise	&	Pixelwise	&	Patchwise	\\
\hline \hline 								Healthy Grass	&	83.29 	&	84.24 	&	82.24 	&	83.67 	&	82.62 	&	82.81 	&	82.53 	&	93.83 	&	\bf{95.92}	\\
Stressed Grass	&	59.59 	&	68.05 	&	86.84 	&	97.74 	&	98.31 	&	98.68 	&	96.71 	&	95.30 	&	\bf{98.87}	\\
Synthetic Grass	&	98.81 	&	91.09 	&	96.24 	&	\bf{100.00}	&	96.44 	&	99.60 	&	98.22 	&	99.80 	&	98.61 	\\
Tree	&	83.71 	&	80.59 	&	\bf{99.72}	&	98.20 	&	98.96 	&	98.48 	&	98.20 	&	96.50 	&	99.05 	\\
Soil	&	46.12 	&	49.62 	&	85.51 	&	97.73 	&	97.92 	&	96.02 	&	98.77 	&	98.86 	&	\bf{99.91}	\\
Water	&	93.71 	&	92.31 	&	\bf{95.80}	&	95.10 	&	95.10 	&	95.10 	&	93.01 	&	95.10 	&	94.41 	\\
Residential	&	59.51 	&	78.64 	&	78.54 	&	86.75 	&	85.26 	&	85.91 	&	\bf{91.32}	&	82.46 	&	87.03 	\\
Commercial	&	69.42 	&	64.01 	&	\bf{82.24}	&	60.21 	&	66.29 	&	50.71 	&	67.43 	&	77.30 	&	64.86 	\\
Road	&	76.30 	&	78.56 	&	70.73 	&	74.22 	&	71.48 	&	73.56 	&	71.95 	&	\bf{82.06}	&	80.64 	\\
Highway	&	53.38 	&	39.77 	&	50.10 	&	81.37 	&	69.40 	&	88.51 	&	86.29 	&	60.33 	&	\bf{98.17}	\\
Railway	&	81.69 	&	56.45 	&	85.77 	&	\bf{88.80}	&	83.02 	&	75.81 	&	81.02 	&	81.40 	&	80.36 	\\
Parking Lot 1	&	52.83 	&	42.94 	&	62.73 	&	68.01 	&	57.44 	&	65.03 	&	65.90 	&	74.26 	&	\bf{81.56}	\\
Parking Lot 2	&	67.02 	&	57.89 	&	81.40 	&	75.79 	&	\bf{84.91}	&	72.63 	&	69.82 	&	72.63 	&	79.65 	\\
Tennis Court	&	92.31 	&	94.33 	&	99.60 	&	99.60 	&	98.79 	&	99.60 	&	97.17 	&	99.19 	&	\bf{100.00}	\\
Running Track	&	96.19 	&	98.10 	&	94.50 	&	97.89 	&	95.35 	&	98.73 	&	\bf{99.15}	&	98.31 	&	98.52 	\\
\hline\hline	
OA (\%) $\uparrow$	&	69.94 	&	67.58 	&	80.57 	&	85.21 	&	82.94 	&	83.32 	&	85.27 	&	\underline{85.64}	&	\bf{89.52}	\\
AA (\%) $\uparrow$	&	74.26 	&	71.77 	&	83.46 	&	87.01 	&	85.42 	&	85.41 	&	86.50 	&	\underline{87.16}	&	\bf{90.50}	\\
$\kappa$ $\uparrow$	&	0.6749 	&	0.6501 	&	0.7894 	&	0.8397 	&	0.8149 	&	0.8193 	&	0.8402 	&	\underline{0.8442}	&	\bf{0.8864}	\\
\hline																			
MACs (M) $\downarrow$	&	\bf{19.64}	&	26.83 	&	640.09 	&	548.45 	&	549.95 	&	667.04 	&	681.19 	&	\underline{23.52}	&	36.21 	\\
Params (K) $\downarrow$	&	305.94 	&	418.77 	&	1192.96 	&	350.61 	&	353.49 	&	72.56 	&	74.10 	&	\bf{14.23}	&	\underline{36.55}	\\

\hlinew{1pt}
\end{tabular}
% }
\label{tab:HU}
\end{table*}

\subsubsection{Longkou HS Dataset}
The third Longkou dataset, also known as WHU-Hi-LongKou, is a specific UAV-borne HS dataset acquired in Longkou Town, Hubei, China. 
This dataset was captured using the Headwall Nano-Hyperspec sensor mounted on a UAV at a flight altitude of 500 meters. 
We use the opened data that has 550 $\times$ 400 pixels at a GSD of 0.463 m, and 270 spectral bands ranging from 400 nm to 1000 nm.
More than 204 thousand labelings with 6 crop types and 3 LULC categories were used for evaluation.
This dataset is part of the WHU-Hi dataset collection, which also includes other two datasets HanChuan and HongHu \cite{zhong2020whu}. 

\subsubsection{Augsburg HS Dataset}
The second HS dataset Augsburg consists of HS imagery acquired by the HySpex -- an airborne imaging spectrometer system operated by the the Remote Sensing Technology Institute of the German Aerospace Center \cite{hong2021multimodal}.
The pre-processed HS imagery has 180 bands ranging from 400 nm to 2500 nm with high quality.
Our selected sub-region includes 332 $\times$ 485 pixels at a GSD of 30m, covering the city of Augsburg, Germany.
The ground reference map comprising 7 categories was elaborately produced by manual labeling based on OpenStreetMap product.

\subsubsection{Botswana HS Dataset}
The last Botswana dataset is a collection of HS imagery acquired by the NASA EO-1 satellite over the Okavango Delta area of Botswana between 2001 and 2004.
This dataset was captured using the well-known Hyperion sensor, which collects original data at a 30-meter spatial resolution in 242 bands covering the 400-2500 nm portion of the spectrum. 
After pre-processing by removing uncalibrated and noisy bands, 145 bands are commonly used for classification experiments. 
The ground reference comprises 14 identified classes representing different land cover types in the investigated region.

\subsection{Experimental Setup}
To ensure the reproduction of our experimental results and a faithful verification of the effectiveness of our method, we herein present the necessary details in implementing all compared methods on investigated datasets.

\begin{figure*}[!t]
    \centering
		\includegraphics[width=0.98\textwidth]{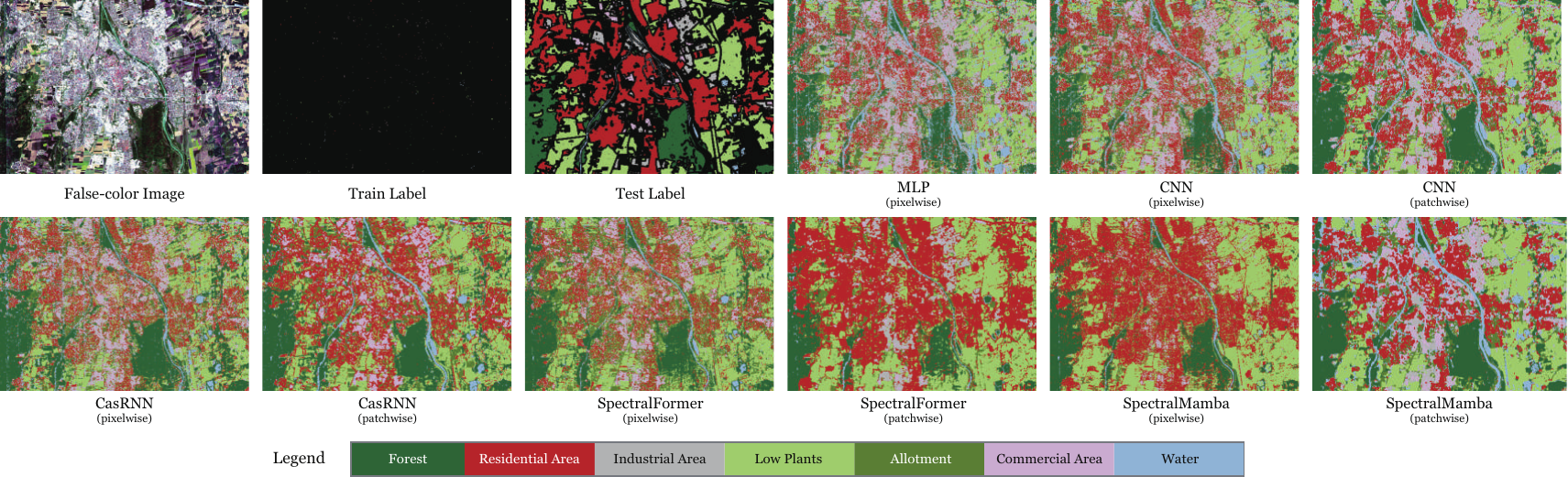}
    \caption{An illustration of the false-color image, train and test labels, and classification maps obtained by compared methods on the Augsburg HS dataset.}
    \label{fig:AG}
\end{figure*}

\begin{table*}[!t]
\scriptsize
\centering
\caption{Quantitative comparison with related representatives on Augsburg HS dataset. 
The best and second-best overall results are shown in bold and underlined, respectively.}
% \resizebox{0.98\textwidth}{!}{ 
\begin{tabular}{c||c|cc|cc|cc|cc}
\hlinew{1pt}		
Method	&	MLP\cite{hong2023cross}	&	\multicolumn{2}{c|}{CNN\cite{yao2022semi}}			&	\multicolumn{2}{c|}{CasRNN\cite{hang2019cascaded}}			&	\multicolumn{2}{c|}{SpectralFormer\cite{hong2022spectralformer}}			&	\multicolumn{2}{c}{SpectralMamba}			\\
\hline																			
Implementation	&	Pixelwise	&	Pixelwise	&	Patchwise	&	Pixelwise	&	Patchwise	&	Pixelwise	&	Patchwise	&	Pixelwise	&	Patchwise	\\
\hline \hline 								Forest	&	13.27 	&	12.15 	&	53.77 	&	94.30 	&	96.31 	&	90.23 	&	92.74 	&	87.96 	&	\bf{98.31}	\\
Residential Area	&	23.61 	&	32.71 	&	35.84 	&	47.53 	&	63.76 	&	45.50 	&	53.58 	&	57.33 	&	\bf{73.79}	\\
Industrial Area	&	34.37 	&	25.40 	&	32.67 	&	30.76 	&	9.71 	&	24.85 	&	\bf{54.65}	&	35.38 	&	48.71 	\\
Low Plants	&	22.90 	&	29.86 	&	38.45 	&	57.23 	&	66.52 	&	71.43 	&	\bf{84.46}	&	77.39 	&	77.60 	\\
Allotment	&	62.63 	&	48.48 	&	80.00 	&	71.11 	&	90.30 	&	82.22 	&	75.35 	&	77.37 	&	\bf{94.14}	\\
Commercial Area	&	42.56 	&	30.22 	&	45.30 	&	45.30 	&	\bf{58.34}	&	53.04 	&	36.17 	&	48.12 	&	58.21 	\\
Water	&	51.83 	&	50.38 	&	61.08 	&	53.42 	&	63.35 	&	49.55 	&	64.67 	&	61.90 	&	\bf{71.98}	\\
\hline\hline																			
OA (\%) $\uparrow$	&	23.26 	&	28.20 	&	40.63 	&	58.35 	&	67.77 	&	61.62 	&	\underline{71.03}	&	68.49 	&	\bf{77.90}	\\
AA (\%) $\uparrow$	&	35.88 	&	32.75 	&	49.59 	&	57.09 	&	64.04 	&	59.54 	&	\underline{65.95}	&	63.64 	&	\bf{74.68}	\\
$\kappa$ $\uparrow$	&	0.0942	&	0.1094	&	0.2692	&	0.4645	&	0.5673	&	0.5018 	&	\underline{0.6157}	&	0.5864	&	\bf{0.7048}	\\
\hline																			
MACs (M) $\downarrow$	&	\bf{20.10}	&	41.74 	&	1038.34 	&	685.38 	&	687.24 	&	832.60 	&	850.29 	&	\underline{30.96}	&	50.55 	\\
Params (K) $\downarrow$	&	313.10 	&	651.61 	&	1861.63 	&	348.55 	&	352.15 	&	72.04 	&	73.58 	&	\bf{13.56}	&	\underline{47.94}	\\

\hlinew{1pt}
\end{tabular}
% }
\label{tab:AG}
\end{table*}

\subsubsection{Train and Test Set Split}
Besides the data quality, the split of the train and test set has a significant effect on assessing the model performance, particularly the deep models.
We strive to reason the popularity of the Houston2013 dataset and conclude four criteria for sample selection with easy practicality and wide acceptability.
\begin{enumerate}
    \item First, the samples selected for training are better evenly distributed in the scene to ameliorate the spectral variability phenomenon that may deteriorate most models.
    \item Second, it is important to maintain a moderate train set size to prevent the evaluation that is either too simplistic or too challenging, therefore striking a balance in accurately reflecting the model's performance and generalization ability.
    \item Third, rather than pixelwise sampling, image segment-based sampling or labeling tends to be more efficient in progressively constructing the train set to a preset size.
    \item Last but not least, class-balanced sampling is always preferable, especially for unseen scenes with no prior knowledge of the class distribution.
\end{enumerate}

Based on the above criteria, we propose the following steps to determine the train and test set for those datasets without benchmark split.
The first step is to apply the SLIC superpixel method to over-segment the image into a large number of segments \cite{achanta2012slic}, most of which can then be made of the same ground object.
The next step is straightforward to randomly collect those homogeneous segments progressively until the budget of training samples for each class is reached.
The classwise budgets for the last three HS datasets we adopt are empirically set to 80, 50, and 20, respectively.
This method can also be used to organize a separate validation set \cite{yao2023ucsl}.

\subsubsection{Evaluation Metrics}
Besides the conventional performance metrics such as classwise accuracy (CA), overall accuracy (OA), average accuracy (AA), and Kappa coefficient ($\kappa$), we introduce two more metrics to evaluate the efficacy of different methods, which are multiply-accumulate operations (MACs) and parameters (Params) of each network.
As the name suggests, MACs refer to the number of multiply-accumulate operations during actual network training.
In our experiments, we set the batch size for all methods as 64 to fairly compute and compare their MACs for one batch.
The less value the MACs and Params take, the less computational resource the corresponding model costs.

\begin{figure*}[!t]
    \centering
		\includegraphics[width=0.98\textwidth]{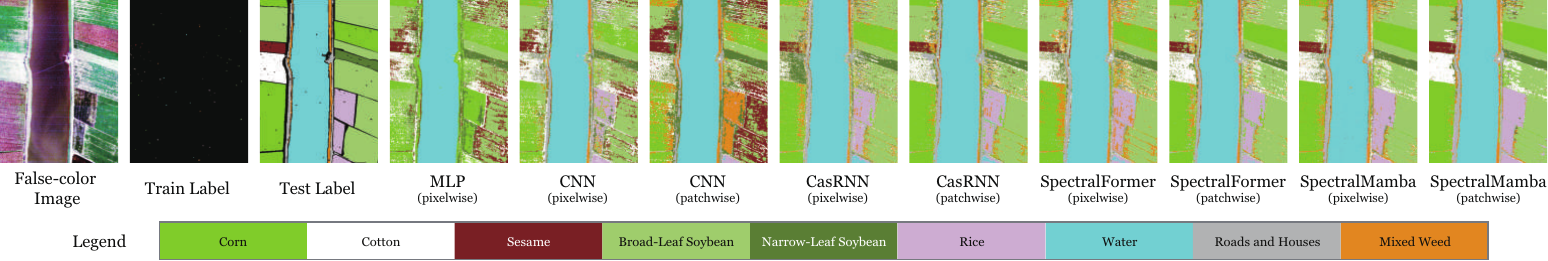}
    \caption{An illustration of the false-color image, train and test labels, and classification maps obtained by compared methods on the Longkou HS dataset.}
    \label{fig:LK}
\end{figure*}

\begin{table*}[!t]
\scriptsize
\centering
\caption{Quantitative comparison with related representatives on Longkou HS dataset. 
The best and second-best overall results are shown in bold and underlined, respectively.}
% \resizebox{0.98\textwidth}{!}{
\begin{tabular}{c||c|cc|cc|cc|cc}
\hlinew{1pt}																			
Method	&	MLP\cite{hong2023cross}	&	\multicolumn{2}{c|}{CNN\cite{yao2022semi}}			&	\multicolumn{2}{c|}{CasRNN\cite{hang2019cascaded}}			&	\multicolumn{2}{c|}{SpectralFormer\cite{hong2022spectralformer}}			&	\multicolumn{2}{c}{SpectralMamba}			\\
\hline		
Implementation	&	Pixelwise	&	Pixelwise	&	Patchwise	&	Pixelwise	&	Patchwise	&	Pixelwise	&	Patchwise	&	Pixelwise	&	Patchwise	\\
\hline \hline 								Corn	&	37.36 	&	12.68 	&	4.64 	&	89.38 	&	95.98 	&	85.47 	&	85.47 	&	91.60 	&	\bf{97.48}	\\
Cotton	&	28.16 	&	25.12 	&	33.29 	&	30.50 	&	70.46 	&	69.53 	&	69.53 	&	74.49 	&	\bf{95.01}	\\
Sesame	&	29.63 	&	30.50 	&	39.13 	&	51.63 	&	65.91 	&	60.36 	&	60.36 	&	78.17 	&	\bf{92.17}	\\
Broad-Leaf Soybean	&	30.95 	&	32.32 	&	47.81 	&	73.85 	&	77.15 	&	73.44 	&	73.44 	&	79.27 	&	\bf{82.87}	\\
Narrow-Leaf Soybean	&	32.63 	&	25.24 	&	66.49 	&	76.42 	&	86.45 	&	83.26 	&	83.26 	&	\bf{90.85}	&	90.68 	\\
Rice	&	0.00 	&	22.29 	&	13.20 	&	80.77 	&	83.62 	&	79.67 	&	79.67 	&	95.41 	&	\bf{98.29}	\\
Water	&	7.47 	&	6.17 	&	80.98 	&	99.83 	&	\bf{99.99}	&	\bf{99.99}	&	\bf{99.99}	&	98.66 	&	99.60 	\\
Roads and Houses	&	0.00 	&	13.53 	&	7.27 	&	79.86 	&	86.17 	&	77.90 	&	77.90 	&	61.76 	&	\bf{88.81}	\\
Mixed Weed	&	0.00 	&	23.99 	&	33.45 	&	48.05 	&	54.63 	&	53.13 	&	53.13 	&	68.75 	&	\bf{73.74}	\\
\hline\hline																			
OA (\%) $\uparrow$	&	20.57 	&	18.51 	&	47.30 	&	82.92 	&	87.69 	&	84.03 	&	\underline{88.86}	&	87.80 	&	\bf{92.48}	\\
AA (\%) $\uparrow$	&	18.47 	&	21.32 	&	36.25 	&	70.03 	&	80.04 	&	75.86 	&	\underline{82.29}	&	82.11 	&	\bf{90.96}	\\
$\kappa$ $\uparrow$	&	0.0575 	&	0.0693 	&	0.3558 	&	0.7815 	&	0.8417 	&	0.7964 	&	\underline{0.8566}	&	0.8432 	&	\bf{0.9028}	\\
\hline																			
MACs (M) $\downarrow$	&	\bf{21.61}	&	93.74 	&	2333.70 	&	954.68 	&	957.48 	&	1276.52 	&	1303.65 	&	\underline{52.79}	&	93.18 	\\
Params (K) $\downarrow$	&	336.65 	&	1499.14 	&	4186.11 	&	349.07 	&	354.47 	&	\underline{72.17}	&	73.71 	&	\bf{18.68}	&	94.55 	\\

\hlinew{1pt}								
\end{tabular}
% }
\label{tab:LK}
\end{table*}

\subsubsection{Compared Methods}
The major aim of our experiments is to assess whether the proposed SpectralMamba can be deemed as a revolutionized DL tool for the HS image classification task.
Therefore, we select four prevalent types of DL-based solutions for comparison, which are MLP, CNN, RNN-based, and Transformer-based models.
The details of our competitors are as follows.

\begin{enumerate}
    \item The MLP contains an input block, two hidden blocks, and a classification head. 
    Each of the first three blocks consists of a fully-connected (FC) layer, a 1-D batch normalization (BN) layer, and a ReLU activation \cite{hong2023cross}.
    \item The CNN contains two convolution blocks and a classification head. 
    The convolution block consists of 1-D or 2-D convolution layers depending on pixelwise or patchwise input \cite{yao2022semi}, corresponding 1-D or 2-D BN layers, and ReLU.
    \item The CasRNN is selected to represent the RNN-based models \cite{hang2019cascaded}.
    It has pixelwise and patchwise versions.
    The pixelwise CasRNN mainly contains cascaded gated recurrent units, while the patchwise one additionally introduces two separable convolution blocks and a max-pooling layer before the pixelwise CasRNN\footnote{\url{https://github.com/RenlongHang/CasRNN}}. 
    \item The pixelwise SpectralFormer and patchwise SpectralFormer are selected to represent the Transformer-based models \cite{hong2022spectralformer}.
    The former one is equipped with the groupwise spectral embedding to enhance local spectral details while the latter adds an FC layer to encode the spatial information from flattened image patches\footnote{\url{https://github.com/danfenghong/IEEE_TGRS_SpectralFormer}}.
\end{enumerate}

\subsubsection{Implementation Details}
All of our experiments are conducted mainly based on the PyTorch framework using a workstation with an Nvidia GeForce GTX 3080 GPU card.
We tuned the learning rate and weight decay hyperparameters via a rough grid search on intervals of $\{10^{-4},5\times10^{-4},10^{-3},5\times10^{-3}\}$.
The epoch is set to 500.
A StepLR scheduler is adopted to shrink the learning rate by multiplying a factor of 0.9 for each 20 epochs.

Another commonly overlooked issue is information leakage, that is, more testing samples will be seen as the patch size of training samples increases \cite{audebert2019deep}.
To mitigate the effect caused by this issue, we set the patch size for all patchwise models as 3.
We claim that this setup is enough for assessment from the perspective of tendency, which also consumes less energy and is aligned with sustainable development goals.

\begin{figure*}[!t]
    \centering
		\includegraphics[width=0.99\textwidth]{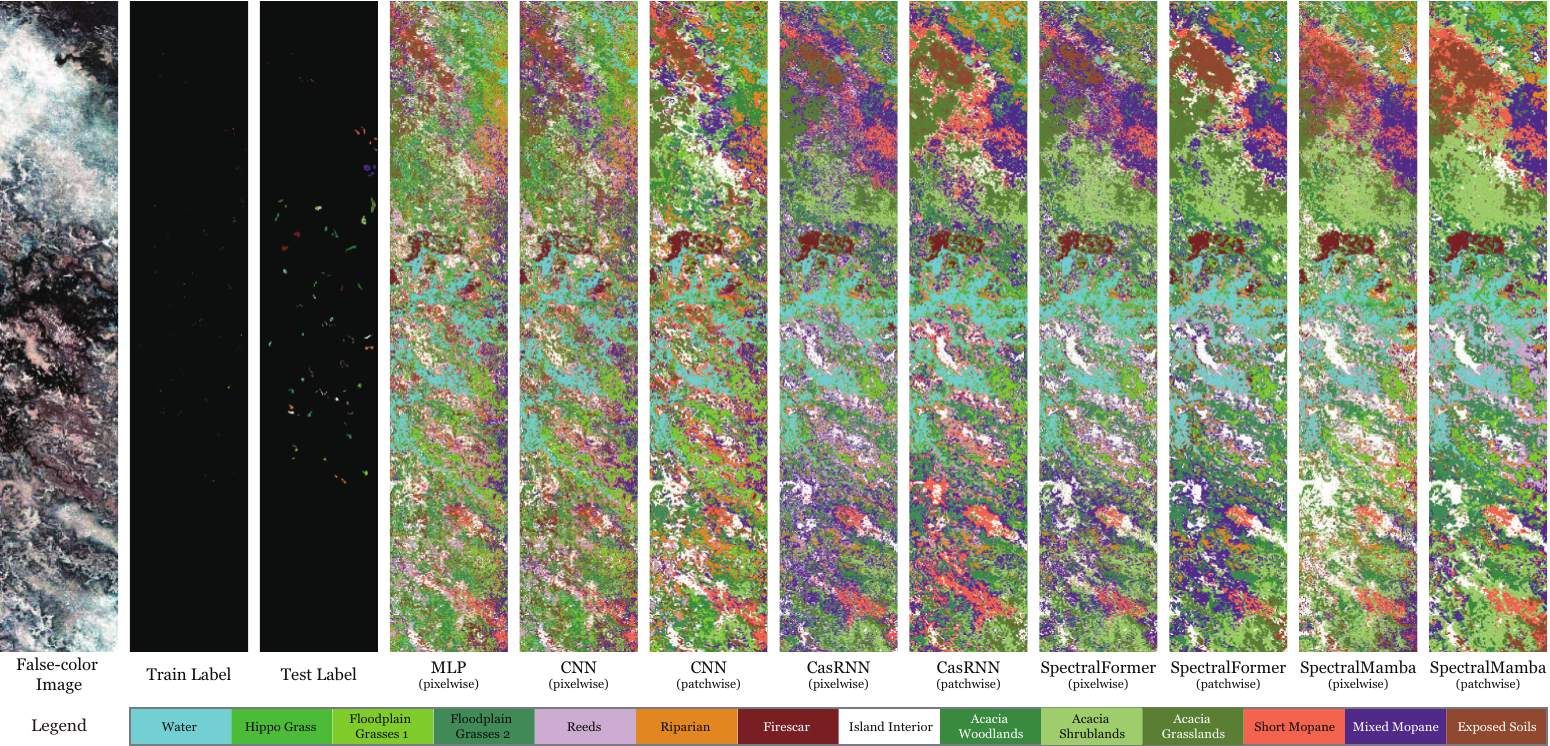}
    \caption{An illustration of the false-color image, train and test labels, and classification maps obtained by compared methods on the Botswana HS dataset.}
    \label{fig:BW}
\end{figure*}

\begin{table*}[!t]
\scriptsize
\centering
\caption{Quantitative comparison with related representatives on Botswana HS dataset. 
The best and second-best overall results are shown in bold and underlined, respectively.}
% \resizebox{0.98\textwidth}{!}{ 
\begin{tabular}{c||c|cc|cc|cc|cc}
\hlinew{1pt}					
Method	&	MLP\cite{hong2023cross}	&	\multicolumn{2}{c|}{CNN\cite{yao2022semi}}			&	\multicolumn{2}{c|}{CasRNN\cite{hang2019cascaded}}			&	\multicolumn{2}{c|}{SpectralFormer\cite{hong2022spectralformer}}			&	\multicolumn{2}{c}{SpectralMamba}			\\
\hline																			
Implementation	&	Pixelwise	&	Pixelwise	&	Patchwise	&	Pixelwise	&	Patchwise	&	Pixelwise	&	Patchwise	&	Pixelwise	&	Patchwise	\\
\hline \hline 								Water	&	\bf{100.00}	&	\bf{100.00}	&	\bf{100.00}	&	\bf{100.00}	&	\bf{100.00}	&	\bf{100.00}	&	\bf{100.00}	&	\bf{100.00}	&	\bf{100.00}	\\
Hippo Grass	&	\bf{100.00}	&	\bf{100.00}	&	\bf{100.00}	&	97.53 	&	97.53 	&	\bf{100.00}	&	98.77 	&	95.06 	&	\bf{100.00}	\\
Floodplain Grasses 1	&	94.78 	&	98.26 	&	\bf{100.00}	&	81.30 	&	96.52 	&	90.87 	&	95.65 	&	95.65 	&	\bf{100.00}	\\
Floodplain Grasses 2	&	96.86 	&	96.86 	&	94.76 	&	88.48 	&	94.76 	&	93.72 	&	92.15 	&	92.15 	&	\bf{97.91}	\\
Reeds	&	90.36 	&	89.16 	&	90.76 	&	75.90 	&	73.49 	&	79.52 	&	87.55 	&	86.75 	&	\bf{93.17}	\\
Riparian	&	71.89 	&	82.73 	&	\bf{97.59}	&	71.08 	&	68.27 	&	65.86 	&	87.15 	&	77.91 	&	94.78 	\\
Firescar	&	97.49 	&	96.23 	&	\bf{100.00}	&	96.65 	&	89.96 	&	87.87 	&	99.58 	&	97.91 	&	\bf{100.00}	\\
Island Interior	&	99.45 	&	98.91 	&	\bf{100.00}	&	91.26 	&	97.81 	&	97.27 	&	\bf{100.00}	&	95.63 	&	\bf{100.00}	\\
Acacia Woodlands	&	84.98 	&	74.74 	&	90.10 	&	72.35 	&	72.35 	&	74.74 	&	87.71 	&	86.35 	&	\bf{96.25}	\\
Acacia Shrublands	&	95.13 	&	93.36 	&	\bf{97.79}	&	73.45 	&	74.34 	&	92.48 	&	95.13 	&	96.02 	&	\bf{97.79}	\\
Acacia Grasslands	&	87.37 	&	92.28 	&	\bf{98.25}	&	85.96 	&	92.28 	&	84.91 	&	97.19 	&	85.61 	&	94.04 	\\
Short Mopane	&	96.23 	&	95.60 	&	\bf{100.00}	&	99.37 	&	\bf{100.00}	&	96.86 	&	\bf{100.00}	&	91.19 	&	99.37 	\\
Mixed Mopane	&	89.52 	&	91.94 	&	98.39 	&	76.61 	&	89.92 	&	88.71 	&	93.55 	&	97.58 	&	\bf{99.60}	\\
Exposed Soils	&	\bf{100.00}	&	98.67 	&	98.67 	&	94.67 	&	98.67 	&	98.67 	&	\bf{100.00}	&	\bf{100.00}	&	\bf{100.00}	\\
\hline\hline																			
OA (\%) $\uparrow$	&	91.81 	&	92.21 	&	\underline{97.19}	&	84.19 	&	87.14 	&	87.44 	&	94.55 	&	91.88 	&	\bf{97.66}	\\
AA (\%) $\uparrow$	&	93.15 	&	93.48 	&	\underline{97.59}	&	86.05 	&	88.99 	&	89.39 	&	95.32 	&	92.70 	&	\bf{98.06}	\\
$\kappa$ $\uparrow$	&	0.9112 	&	0.9156 	&	\underline{0.9695}	&	0.8287 	&	0.8606 	&	0.8639 	&	0.9409 	&	0.9119 	&	\bf{0.9747}	\\
\hline																			
MACs (M) $\downarrow$	&	\bf{19.64}	&	27.19 	&	658.12 	&	570.80 	&	572.31 	&	671.63 	&	685.89 	&	\underline{23.39}	&	36.25 	\\
Params (K) $\downarrow$	&	305.93 	&	424.43 	&	1209.34 	&	350.35 	&	353.25 	&	72.49 	&	74.03 	&	\bf{12.33}	&	\underline{34.96}	\\

\hlinew{1pt}
\end{tabular}
% }
\label{tab:BW}
\end{table*}

\subsection{Results and Analysis}
We summarize the quantitative results of all compared methods on six metrics, i.e., performance metrics as CA, OA, AA, and $\kappa$, and efficacy metrics as MACs and Params, in Tables \ref{tab:HU} to \ref{tab:BW} for Houston2013, Augsburg, Longkou, and Botswana datasets, respectively.
The classification maps of full scenes are correspondingly visualized in Figs. \ref{fig:HU} to \ref{fig:BW}.

From the tables, we can draw several conclusions that are consistent on four datasets.
Patchwise implementations generally tend to produce better performance than pixelwise ones by exploitation of spatial information (except for CasRNN on the Houston2013 dataset, which is probably caused by the unstable training issue).
In specific, pixelwise MLP and pixelwise CNN involve fewer computations than CasRNN and SpectralFormer. 
Still, patchwise CNN encounters an evident explosion in terms of MACs due to its use of inseparable 2-D convolution.
If we examine it further through the lens of computational efficiency, although SpectralFormer owns much fewer parameters than those of CasRNN, its self-attention operations result in higher MACs, particularly for dealing with longer sequences in the middle two datasets.
Most importantly, our SpectralMamba unsurprisingly achieves the best CAs in most classes, the best OA, AA, and $\kappa$, and meanwhile the best Params, and the MACs that are only second to those of the simplistic MLP with a negligible margin.

Some other interesting observations are also worth noting.
If we set the quantitative tendency on the Houston2013 dataset as the reference, both the conventional pixelwise MLP and pixelwise CNN behave contrarily on Botswana and the other two datasets.
This could possibly be explained by that the train and test sample distributions are much closer to each other on the Botswana dataset than on the other two.
Patchwise CNN raises the performance of pixelwise CNN by 12.43$\%$ and 28.79$\%$ in terms of OA on the middle two datasets, but still at a low level.
CasRNN, SpectralFormer, and our SpectralMamba, no matter whether pixelwise or patchwise implementation, consistently show a stable classification performance at a desirable level, which verifies the importance and superiority of their sequential modeling capability.
Furthermore, our SpectralMamba successfully addresses the computational weaknesses that lie in the former two prevalent sequence models by leveraging appropriate spatial-spectral merging, sequential scanning, and state space modeling.

We can draw clues from the visualization results that exhibit trends close to those of the above-mentioned quantitative results.
For example, our SpectralMamba offers more reliable predictions for categories such as \textit{road} and \textit{highway} in the Houston2013 scene.
As for the Augsburg scene, although CasRNN and SpectralFormer also show \textit{forest} as good as ours, they typically confuse \textit{residential area} with other classes.
Moreover, the trajectory of the river and the shape of paddy fields, i.e., \textit{water} category, can be clearly depicted by our SpectralMamba.
The superiority of our methods is more obvious in the Longkou scene.
Take \textit{rice} area as an instance, in contrast to other methods that are easy to mix it with \textit{water} or \textit{mixed weed}, ours discover a region that is most regular and homogeneous used to plant \textit{rice}.
Although the labels in the Botswana scene are more sparse, we can still tell that our patchwise SpectralMamba generates classification maps with less salt and pepper noises-like patterns that are more likely in accord with the real distribution of ground objects.

\begin{table}[!t]
% \scriptsize
\centering
\caption{Quantitative results of ablation by switching on and off three key modules in the proposed SpectralMamba on Houston2013 HS dataset. 
The relative \tcg{improvement} and \tcr{degradation} are shown in \tcg{cyan} and \tcr{red}, respectively.}
\resizebox{0.48\textwidth}{!}{
\begin{tabular}{c||cc|cccc}
\hlinew{1pt}	
\multirow{2}{*}{Module}	&	\multicolumn{6}{c}{Implementation}										\\
\cline{2-7}	
	&	\multicolumn{2}{c|}{Pixelwise}			&	\multicolumn{4}{c}{Patchwise}							\\
\hline\hline
GSSM	&	-	&	-	&	\xmark	&	\cmark	&	\cmark	&	\cmark	\\
PSS	&	\xmark	&	\cmark	&	\cmark	&	\xmark	&	\cmark	&	\cmark	\\
Mamba	&	\cmark	&	\cmark	&	\cmark	&	\cmark	&	\xmark	&	\cmark	\\
\hline\hline		
\multirow{2}{*}{OA (\%) $\uparrow$}	&	\multirow{2}{*}{84.62}	&	85.59	&	86.74	&	85.73	&	84.40	&	89.52	\\
	&		&	\tcg{(+0.97)}	&	\tcg{(+2.12)}	&	\tcg{(+1.11)}	&	\tcr{(-0.22)}	&	\tcg{(+4.90)}	\\
\multirow{2}{*}{AA (\%) $\uparrow$}	&	\multirow{2}{*}{85.56}	&	87.23	&	88.40	&	87.55	&	86.57	&	90.50	\\
	&		&	\tcg{(+1.67)}	&	\tcg{(+2.84)}	&	\tcg{(+1.99)}	&	\tcg{(+1.01)}	&	\tcg{(+4.94)}	\\
\multirow{2}{*}{$\kappa$ $\uparrow$}	&	\multirow{2}{*}{0.8330}	&	0.8438	&	0.8561	&	0.8451	&	0.8308	&	0.8864	\\
	&		&	\tcg{(+0.0108)}	&	\tcg{(+0.0231)}	&	\tcg{(+0.0121)}	&	\tcr{(-0.0008)}	&	\tcg{(+0.0534)}	\\
\hline													
\multirow{2}{*}{MACs (M) $\downarrow$}	&	\multirow{2}{*}{43.21}	&	23.25	&	23.60	&	55.90	&	12.73	&	36.21	\\
	&		&	\tcg{(-19.96)}	&	\tcg{(-19.61)}	&	\tcr{(+12.69)}	&	\tcg{(-30.48)}	&	\tcg{(-7.00)}	\\
\multirow{2}{*}{Params (K) $\downarrow$}	&	\multirow{2}{*}{66.17}	&	14.23	&	14.24	&	88.49	&	22.75	&	36.55	\\
	&		&	\tcg{(-51.94)}	&	\tcg{(-51.93)}	&	\tcr{(+22.32)}	&	\tcg{(-43.42)}	&	\tcg{(-29.62)}	\\
\hlinew{1pt}								
\end{tabular}}
\label{tab:Ablation1}
\end{table}

\subsection{Ablation Studies}
% counterpart

\subsubsection{Module Effectivity}
We first conduct ablation experiments on three key components in our SpectralMamba, that are, GSSM, PSS, and Mamba.
We make a quantitative comparison by removing one or two constituents are summarized in Table \ref{tab:Ablation1}.
In most cases, an evident relative improvement can be observed on the basis of a single use of Mamba under pixelwise implementation.
For example, our proposed scanning strategy brings nearly 1$\%$ and 3$\%$ accuracy improvement under pixelwise and patchwise settings, respectively.
What's more, it can lead to 20M and 50K decreases in MACs and Params respectively, compared to the scenarios where only this module is switched off.
We also observe a significant improvement in terms of 2.78$\%$ OA and 2.10$\%$ AA by activating the GSSM module in patchwise case.
Note that the Mamba can still be deemed the most valuable module as more than 5$\%$ OA decrease occurs when removing it.

\subsubsection{Scanning Pieces}
\begin{figure}[!t]
    \centering
		\includegraphics[width=0.48\textwidth]{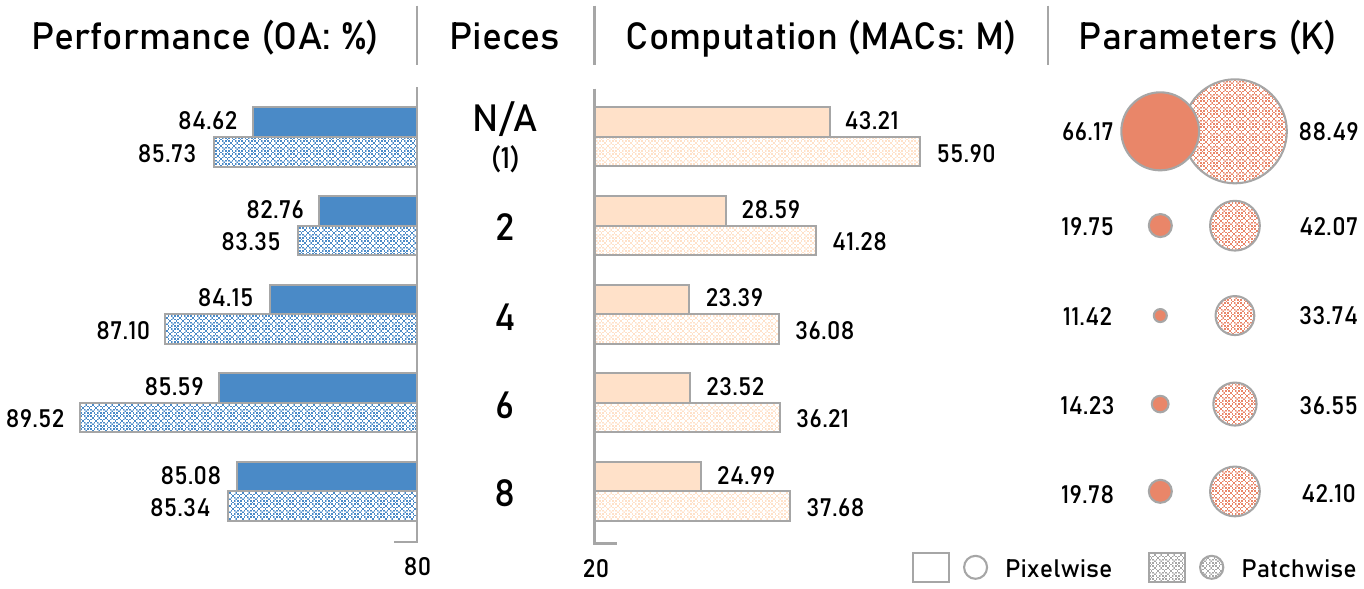}
    \caption{Ablation results of our SpectralMamba on Houston2013 HS Dataset by varying the number of pieces for the proposed piece-wise sequential scanning.}
    \label{fig:Abalation2}
\end{figure}

The other ablation experiment worth conducting is to investigate the impact of varying the number of scanning pieces, i.e., $R$.
The quantitative results are illustrated in Fig. \ref{fig:Abalation2} by setting $R$ to values from 2 to 8 with an interval of 2.
From the figure, we can observe that the baseline without using a scanning strategy can still attain an acceptable performance, while its computation and parameter scales are much higher than those using scanning.
Specifically, in both pixelwise and patchwise cases, the MACs and Params metrics simultaneously show an initial decrease followed by an increase as $R$ increases, in contrast to the trend for performance.
The underlying reason behind this phenomenon lies in that too many pieces inevitably widen the networks for SSM and increase the difficulty in pattern recognition from resulting spectral sequences with short lengths.
Fortunately, by trading off all metrics, we can find the optimal $R=6$ for the Houston2013 dataset, whose computation and parameters are maximally 46$\%$ and 79$\%$ less than w/o scanning while bringing 1$\%$ to 4$\%$ improvements in OA approximately.

\section{Conclusion}\label{S4}
The issues of spectral redundancy and spectral variability have long been plaguing humans from precisely sensing and perceiving the world via HS imaging.
Existing intelligent methods for solving the HS image classification task, based on either conventional CNNs or trending sequence models like RNNs and Transformers, can hardly address these issues simultaneously and efficiently.
Building upon the foundations laid by recent advances in SSMs, we establish the first HS data-oriented deep SSM model, i.e., SpectralMamba, by elaborately proposing PSS and GSSM to ease the sequentiality learning in the state domain and rectify the spectrum in the spatial-spectral domain, respectively.
SpectralMamba enjoys a surprisingly simplified and lightweight architecture and achieves credibly superior HS image classification performance in most aspects.
In the future, we will endeavor to unveil its potential for more critical applications using HS data under resource-constrained scenarios.

% \section*{Acknowledgments}
% This should be a simple paragraph before the References to thank those individuals and institutions who have supported your work on this article.

%% Abstract

% , the highly spatial-spectral-variant HS data dynamics are adequately and efficiently captured 
% Featuring the adequate and simplified modeling of HS data dynamics simultaneously in spatial-spectral and hidden state spaces, SpectralMamba surprisingly creates win-wins from both efficacy and efficiency perspectives.
% Specifically, by ably introducing the latest selective SMM
% unravel the hundreds-of-bands spectrum into pieces
% one of the non-ignorable issues still has found its lies in their computational inefficiency in either training phase or inference.

% {\appendix[Proof of the Zonklar Equations]
% Use $\backslash${\tt{appendix}} if you have a single appendix:
% Do not use $\backslash${\tt{section}} anymore after $\backslash${\tt{appendix}}, only $\backslash${\tt{section*}}.
% If you have multiple appendixes use $\backslash${\tt{appendices}} then use $\backslash${\tt{section}} to start each appendix.
% You must declare a $\backslash${\tt{section}} before using any $\backslash${\tt{subsection}} or using $\backslash${\tt{label}} ($\backslash${\tt{appendices}} by itself
%  starts a section numbered zero.)}


% \bibliographystyle{IEEEtran}
\bibliographystyle{ieeetr}

\bibliography{ref}

\begin{thebibliography}{10}

\bibitem{bioucas2013hyperspectral}
J.~M. Bioucas-Dias, A.~Plaza, G.~Camps-Valls, P.~Scheunders, N.~Nasrabadi, and J.~Chanussot, ``Hyperspectral remote sensing data analysis and future challenges,'' {\em IEEE Geoscience and Remote Sensing Magazine}, vol.~1, no.~2, pp.~6--36, 2013.

\bibitem{xi2022few}
B.~Xi, J.~Li, Y.~Li, R.~Song, D.~Hong, and J.~Chanussot, ``Few-shot learning with class-covariance metric for hyperspectral image classification,'' {\em IEEE Transactions on Image Processing}, vol.~31, pp.~5079--5092, 2022.

\bibitem{hong2024spectralgpt}
D.~Hong, B.~Zhang, X.~Li, Y.~Li, C.~Li, J.~Yao, N.~Yokoya, H.~Li, P.~Ghamisi, X.~Jia, A.~Plaza, P.~Gamba, J.~A. Benediktsson, and J.~Chanussot, ``Spectralgpt: Spectral remote sensing foundation model,'' {\em IEEE Transactions on Pattern Analysis and Machine Intelligence}, 2024.
\newblock DOI:10.1109/TPAMI.2024.3362475.

\bibitem{chang2023changes}
S.~Chang and P.~Ghamisi, ``Changes to captions: An attentive network for remote sensing change captioning,'' {\em IEEE Transactions on Image Processing}, 2023.

\bibitem{ahmad2022hyperspectral}
M.~Ahmad, S.~Shabbir, S.~K. Roy, D.~Hong, X.~Wu, J.~Yao, A.~M. Khan, M.~Mazzara, S.~Distefano, and J.~Chanussot, ``Hyperspectral image classification—traditional to deep models: A survey for future prospects,'' {\em IEEE Journal of Selected Topics in Applied Earth Observations and Remote Sensing}, vol.~15, pp.~968--999, 2022.

\bibitem{hong2021joint}
D.~Hong, N.~Yokoya, J.~Chanussot, J.~Xu, and X.~X. Zhu, ``Joint and progressive subspace analysis (jpsa) with spatial--spectral manifold alignment for semisupervised hyperspectral dimensionality reduction,'' {\em IEEE Transactions on Cybernetics}, vol.~51, no.~7, pp.~3602--3615, 2021.

\bibitem{ma2013hughes}
W.~Ma, C.~Gong, Y.~Hu, P.~Meng, and F.~Xu, ``The hughes phenomenon in hyperspectral classification based on the ground spectrum of grasslands in the region around qinghai lake,'' in {\em International Symposium on Photoelectronic Detection and Imaging 2013: Imaging Spectrometer Technologies and Applications}, vol.~8910, pp.~363--373, SPIE, 2013.

\bibitem{theiler2019spectral}
J.~Theiler, A.~Ziemann, S.~Matteoli, and M.~Diani, ``Spectral variability of remotely sensed target materials: Causes, models, and strategies for mitigation and robust exploitation,'' {\em IEEE Geoscience and Remote Sensing Magazine}, vol.~7, no.~2, pp.~8--30, 2019.

\bibitem{camps2005kernel}
G.~Camps-Valls and L.~Bruzzone, ``Kernel-based methods for hyperspectral image classification,'' {\em IEEE Transactions on Geoscience and Remote Sensing}, vol.~43, no.~6, pp.~1351--1362, 2005.

\bibitem{camps2013advances}
G.~Camps-Valls, D.~Tuia, L.~Bruzzone, and J.~A. Benediktsson, ``Advances in hyperspectral image classification: Earth monitoring with statistical learning methods,'' {\em IEEE Signal Processing Magazine}, vol.~31, no.~1, pp.~45--54, 2013.

\bibitem{hong2021multimodal}
D.~Hong, J.~Hu, J.~Yao, J.~Chanussot, and X.~X. Zhu, ``Multimodal remote sensing benchmark datasets for land cover classification with a shared and specific feature learning model,'' {\em ISPRS Journal of Photogrammetry and Remote Sensing}, vol.~178, pp.~68--80, 2021.

\bibitem{hong2019learnable}
D.~Hong, N.~Yokoya, N.~Ge, J.~Chanussot, and X.~X. Zhu, ``Learnable manifold alignment (lema): A semi-supervised cross-modality learning framework for land cover and land use classification,'' {\em ISPRS Journal of Photogrammetry and Remote Sensing}, vol.~147, pp.~193--205, 2019.

\bibitem{samat2014rm}
A.~Samat, P.~Du, S.~Liu, J.~Li, and L.~Cheng, ``${{\rm E}^{2}}{\rm lms} $: Ensemble extreme learning machines for hyperspectral image classification,'' {\em IEEE Journal of Selected Topics in Applied Earth Observations and Remote Sensing}, vol.~7, no.~4, pp.~1060--1069, 2014.

\bibitem{haut2018active}
J.~M. Haut, M.~E. Paoletti, J.~Plaza, J.~Li, and A.~Plaza, ``Active learning with convolutional neural networks for hyperspectral image classification using a new bayesian approach,'' {\em IEEE Transactions on Geoscience and Remote Sensing}, vol.~56, no.~11, pp.~6440--6461, 2018.

\bibitem{li2023lrr}
C.~Li, B.~Zhang, D.~Hong, J.~Yao, and J.~Chanussot, ``Lrr-net: An interpretable deep unfolding network for hyperspectral anomaly detection,'' {\em IEEE Transactions on Geoscience and Remote Sensing}, 2023.
\newblock DOI:10.1109/TGRS.2023.3279834.

\bibitem{xu2023txt2img}
Y.~Xu, W.~Yu, P.~Ghamisi, M.~Kopp, and S.~Hochreiter, ``Txt2img-mhn: Remote sensing image generation from text using modern hopfield networks,'' {\em IEEE Transactions on Image Processing}, 2023.

\bibitem{gu2021efficiently}
A.~Gu, K.~Goel, and C.~R{\'e}, ``Efficiently modeling long sequences with structured state spaces,'' {\em International Conference on Learning Representations}, 2022.

\bibitem{kalman1960new}
R.~E. Kalman, ``A new approach to linear filtering and prediction problems,'' 1960.

\bibitem{nguyen2022s4nd}
E.~Nguyen, K.~Goel, A.~Gu, G.~Downs, P.~Shah, T.~Dao, S.~Baccus, and C.~R{\'e}, ``S4nd: Modeling images and videos as multidimensional signals with state spaces,'' {\em Advances in Neural Information Processing Systems}, vol.~35, pp.~2846--2861, 2022.

\bibitem{gu2023mamba}
A.~Gu and T.~Dao, ``Mamba: Linear-time sequence modeling with selective state spaces,'' {\em arXiv preprint arXiv:2312.00752}, 2023.

\bibitem{hu2018squeeze}
J.~Hu, L.~Shen, and G.~Sun, ``Squeeze-and-excitation networks,'' in {\em Proceedings of the IEEE Conference on Computer Vision and Pattern Recognition}, pp.~7132--7141, 2018.

\bibitem{liu2024vmamba}
Y.~Liu, Y.~Tian, Y.~Zhao, H.~Yu, L.~Xie, Y.~Wang, Q.~Ye, and Y.~Liu, ``Vmamba: Visual state space model,'' {\em arXiv preprint arXiv:2401.10166}, 2024.

\bibitem{hang2019cascaded}
R.~Hang, Q.~Liu, D.~Hong, and P.~Ghamisi, ``Cascaded recurrent neural networks for hyperspectral image classification,'' {\em IEEE Transactions on Geoscience and Remote Sensing}, vol.~57, no.~8, pp.~5384--5394, 2019.

\bibitem{hong2022spectralformer}
D.~Hong, Z.~Han, J.~Yao, L.~Gao, B.~Zhang, A.~Plaza, and J.~Chanussot, ``Spectralformer: Rethinking hyperspectral image classification with transformers,'' {\em IEEE Transactions on Geoscience and Remote Sensing}, vol.~60, pp.~1--15, 2022.
\newblock Doi: 10.1109/TGRS.2021.3130716.

\bibitem{yu2019free}
J.~Yu, Z.~Lin, J.~Yang, X.~Shen, X.~Lu, and T.~S. Huang, ``Free-form image inpainting with gated convolution,'' in {\em Proceedings of the IEEE/CVF International Conference on Computer Vision}, pp.~4471--4480, 2019.

\bibitem{yao2023extended}
J.~Yao, B.~Zhang, C.~Li, D.~Hong, and J.~Chanussot, ``Extended vision transformer (exvit) for land use and land cover classification: A multimodal deep learning framework,'' {\em IEEE Transactions on Geoscience and Remote Sensing}, 2023.
\newblock 10.1109/TGRS.2023.3284671.

\bibitem{hong2023cross}
D.~Hong, B.~Zhang, H.~Li, Y.~Li, J.~Yao, C.~Li, M.~Werner, J.~Chanussot, A.~Zipf, and X.~X. Zhu, ``Cross-city matters: A multimodal remote sensing benchmark dataset for cross-city semantic segmentation using high-resolution domain adaptation networks,'' {\em Remote Sensing of Environment}, vol.~299, p.~113856, 2023.

\bibitem{yao2022semi}
J.~Yao, X.~Cao, D.~Hong, X.~Wu, D.~Meng, J.~Chanussot, and Z.~Xu, ``Semi-active convolutional neural networks for hyperspectral image classification,'' {\em IEEE Transactions on Geoscience and Remote Sensing}, vol.~60, pp.~1--15, 2022.
\newblock DOI: 10.1109/TGRS.2022.3206208.

\bibitem{zhong2020whu}
Y.~Zhong, X.~Hu, C.~Luo, X.~Wang, J.~Zhao, and L.~Zhang, ``Whu-hi: Uav-borne hyperspectral with high spatial resolution (h2) benchmark datasets and classifier for precise crop identification based on deep convolutional neural network with crf,'' {\em Remote Sensing of Environment}, vol.~250, p.~112012, 2020.

\bibitem{achanta2012slic}
R.~Achanta, A.~Shaji, K.~Smith, A.~Lucchi, P.~Fua, and S.~S{\"u}sstrunk, ``Slic superpixels compared to state-of-the-art superpixel methods,'' {\em IEEE Transactions on Pattern Analysis and Machine Intelligence}, vol.~34, no.~11, pp.~2274--2282, 2012.

\bibitem{yao2023ucsl}
J.~Yao, D.~Hong, H.~Wang, H.~Liu, and J.~Chanussot, ``Ucsl: Towards unsupervised common subspace learning for cross-modal image classification,'' {\em IEEE Transactions on Geoscience and Remote Sensing}, 2023.
\newblock Doi: 10.1109/TGRS.2023.3282951.

\bibitem{audebert2019deep}
N.~Audebert, B.~Le~Saux, and S.~Lef{\`e}vre, ``Deep learning for classification of hyperspectral data: A comparative review,'' {\em IEEE Geoscience and Remote Sensing Magazine}, vol.~7, no.~2, pp.~159--173, 2019.

\end{thebibliography}

% \vspace{-33pt}
% \end{IEEEbiography}

\vfill

\end{document}